\documentclass[letterpaper, 10 pt, conference]{IEEEtran}  

	\newif\ifshowcomments								
	\showcommentstrue				
	\showcommentsfalse				

\title{
Refined Plane Segmentation for Cuboid-Shaped Objects by Leveraging Edge Detection
}

\author{
	\IEEEauthorblockN{
		Alexander Naumann\IEEEauthorrefmark{1},
		Laura D\"orr\IEEEauthorrefmark{1},
		Niels Ole Salscheider\IEEEauthorrefmark{1},
		Kai Furmans\IEEEauthorrefmark{2},
	}
	\IEEEauthorblockA{
		\IEEEauthorrefmark{1}FZI Research Center for Information Technology,
		Karlsruhe, Germany\\
		Email: \{anaumann, doerr, salscheider\}@fzi.de}
	\IEEEauthorblockA{\IEEEauthorrefmark{2}Institute for Material Handling and Logistics, Karlsruhe Institute of Technology (KIT),
	Karlsruhe, Germany\\
	Email: \{kai.furmans\}@kit.edu}
}

\usepackage[pdftex]{graphicx}
\graphicspath{{figures/}}
\usepackage{hyperref}
\hypersetup{
    colorlinks=true,
    linkcolor=black,
    citecolor=black,
    filecolor=black,
    urlcolor=black,
}
\usepackage[T1]{fontenc}
\usepackage[utf8]{inputenc}
\usepackage{csquotes}
\usepackage[english]{babel}
\usepackage[export]{adjustbox}
\usepackage{caption}
\usepackage{subcaption}
\usepackage{lipsum}

\usepackage{placeins}

    \usepackage{changes} 		

	\usepackage{array,multirow,graphicx}

	\usepackage[ruled,vlined]{algorithm2e}

	\usepackage{float}

	\usepackage{siunitx}
	\usepackage{dcolumn}
	
	\usepackage{pgfplotstable}

\usepackage[acronym]{glossaries}

\usepackage{amsmath}
\usepackage{amsfonts}
\usepackage{amssymb}
\usepackage{graphicx}
\usepackage{amsthm}
\usepackage{enumerate}  
\usepackage{subcaption}

		\usepackage{mathtools}
			
		\usepackage{breqn}
		\usepackage{nicefrac}
	
		\usepackage{hhline}

	\usepackage{pgfplots} 
	\pgfplotsset{compat=1.13}
	\usetikzlibrary{plotmarks}
	\usetikzlibrary{patterns}
	\usepgfplotslibrary{external}  
	\tikzsetexternalprefix{TikZ/}
	
	\usepackage{ifthen}
	
	\newcommand{\norm}[1]{\left\lVert#1\right\rVert}

\definecolor{mygrey}{gray}{0.65}
\ifshowcomments
	
\else
	\usepackage{comment}
\fi

\ifshowcomments
	\newcommand{\pcite}[2]{{\color{gray}(\cite{#1} #2)}}  
	\newcommand{\scomment}[1]{{\color{gray} [#1]}}  
\else
	\newcommand{\pcite}[2]{\iffalse #1, #2 \fi}
	\newcommand{\scomment}[1]{\iffalse #1 \fi}
\fi

\newcommand{\fig}[1]{Figure \ref{#1}}
\newcommand{\secr}[1]{Section \ref{#1}}

	\usepackage[style=ieee,
            doi=false,
            url=false,
            isbn=false,
            mincitenames=1,
            maxcitenames=1,
            minbibnames=6,
            maxbibnames=6,
            backend=biber]{biblatex}  

\DeclareSourcemap{
  \maps[datatype=bibtex, overwrite]{
    \map{
      \step[fieldset=address, null]
      \step[fieldset=location, null]
      \step[fieldset=note, null]
      \step[fieldset=series, null]
      \step[fieldset=issn, null]
      \step[fieldset=pages, null]
      \step[fieldset=editor, null]
    }
  }
}

\newcommand{\citet}[1]{\citetitle{#1}}  

\newglossaryentry{i4.0}{name={Industry 4.0}, description={Industry 4.0}}
\newglossaryentry{forklift}{name={forklift truck}, description={}, plural={forklift trucks}}
\newacronym[longplural={automated guided vehicles}]{agv}{AGV}{automated guided vehicle}

\newacronym[longplural={time-of-flight cameras}]{tofc}{ToF camera}{time-of-flight camera}
\newglossaryentry{kinect}{name={MS Kinect}, description={}}
\newacronym{pmd}{PMD}{Photonic Mixing Device}
\newacronym{roi}{ROI}{Region of Interest}
\newacronym{iou}{IoU}{Intersection over Union}
\newglossaryentry{ar}{name={Augmented Reality}, description={}}
\newglossaryentry{rgbd}{name={RGB-D}, description={}}
\newglossaryentry{opencv}{name={OpenCV}, description={}}
\newacronym[longplural={Light Detection And Ranging}]{lidar}{LiDAR}{Light Detection And Ranging}

\newacronym{ransac}{RANSAC}{Random Sampling Consensus}
\newglossaryentry{sota}{name={state-of-the-art}, description={}}
\newglossaryentry{nn}{name={Neural Network}, description={}, plural={Neural Networks}}
\newglossaryentry{cnn}{name={Convolutional Neural Network}, description={}, plural={Convolutional Neural Networks}}
\newacronym{dbscan}{DBSCAN}{Density-Based Spatial Clustering of Applications with Noise}

\newacronym[longplural={Frames per Second}]{fps}{FPS}{Frame per Second}
\newacronym{eu}{EU}{European Union}
\newglossaryentry{poc}{name={proof of concept}, description={}}
	
\begin{document}
\maketitle
\thispagestyle{empty}
\pagestyle{empty}

	\begin{abstract}
Recent advances in the area of plane segmentation from single RGB images show strong accuracy improvements and now allow a reliable segmentation of indoor scenes into planes. 
Nonetheless, fine-grained details of these segmentation masks are still lacking accuracy, thus restricting the usability of such techniques on a larger scale in numerous applications, such as inpainting for \gls{ar} use cases.
We propose a post-processing algorithm to align the segmented plane masks with edges detected in the image. 
This allows us to increase the accuracy of state-of-the-art approaches, while limiting ourselves to cuboid-shaped objects. 
Our approach is motivated by logistics, where this assumption is valid and refined planes can be used to perform robust object detection without the need for supervised learning.
Results for two baselines and our approach are reported on our own dataset, which we made publicly available.
The results show a consistent improvement over the \gls{sota}.
The influence of the prior segmentation and the edge detection is investigated and finally, areas for future research are proposed.
\end{abstract}

	\section{Introduction}
\label{introduction}

\begin{figure}[t!] 
  \begin{subfigure}[b]{0.5\linewidth}
    \centering
    \includegraphics[width=0.9\linewidth]{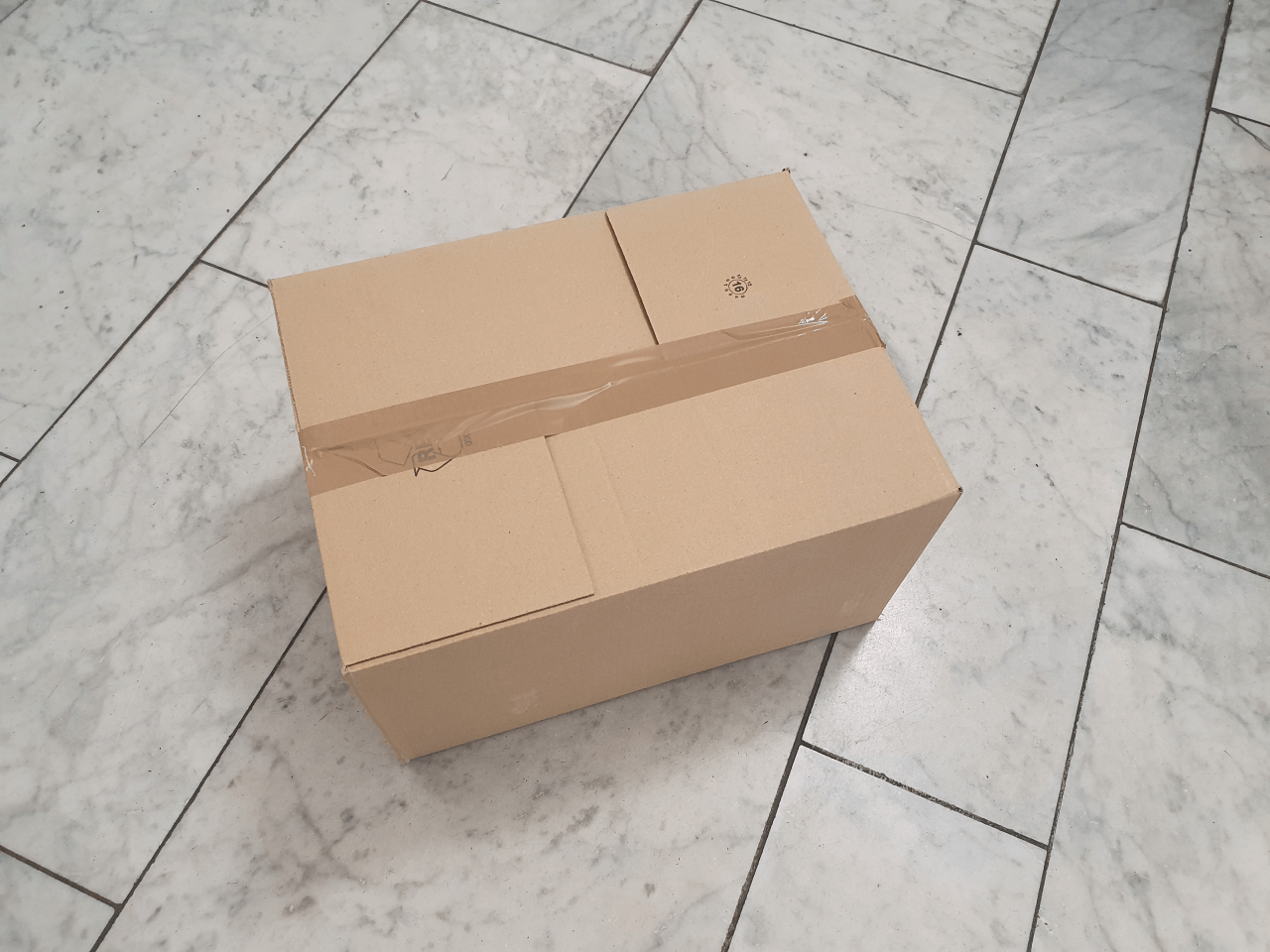} 
    \caption{Input image} 
	\label{fig:overview:a}
    \vspace{2ex}
  \end{subfigure}
  \begin{subfigure}[b]{0.5\linewidth}
    \centering
    \includegraphics[width=0.9\linewidth]{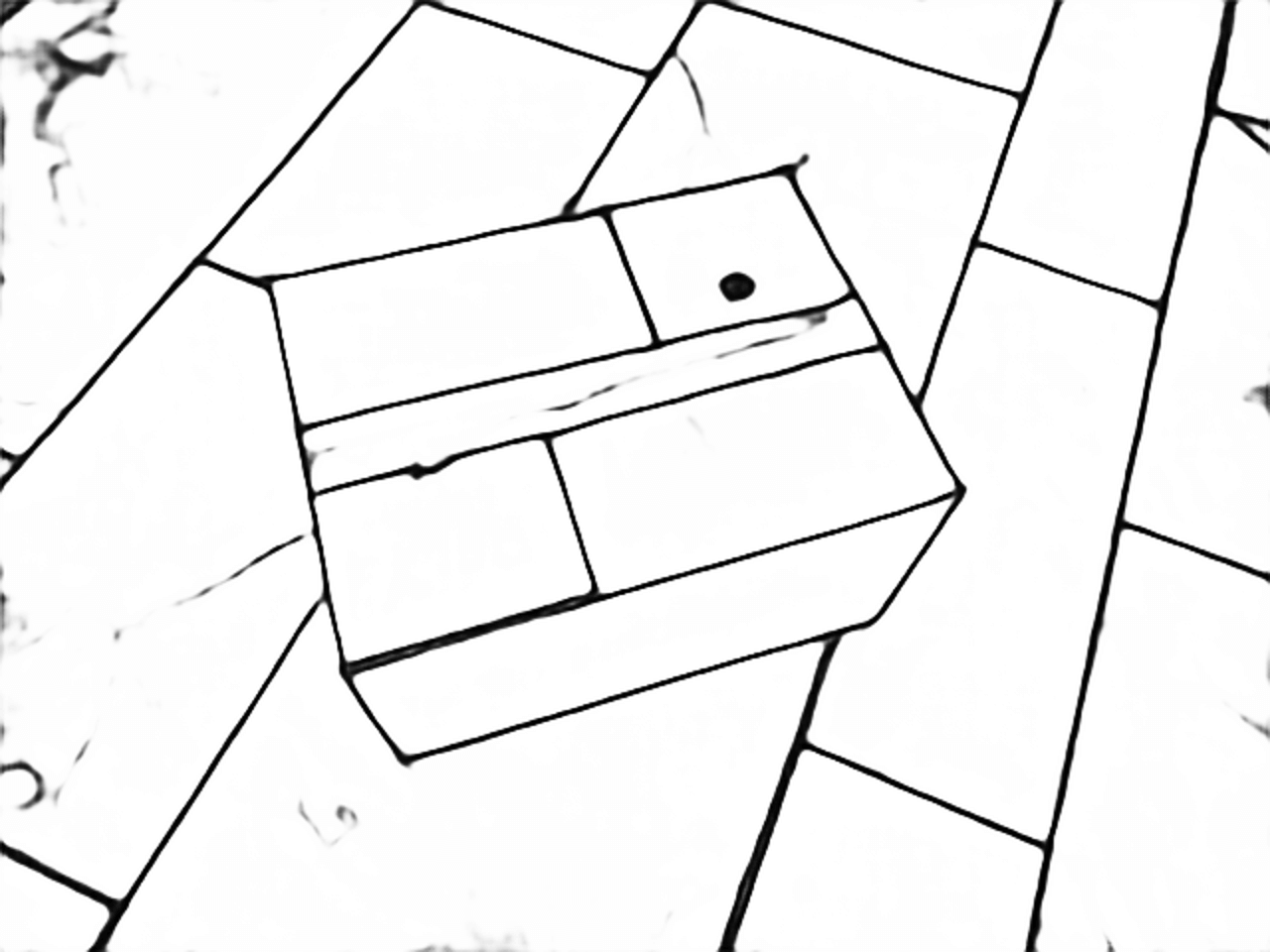} 
    \caption{Edge detection} 
	\label{fig:overview:b}
    \vspace{2ex}
  \end{subfigure} 
  \begin{subfigure}[b]{0.5\linewidth}
    \centering
    \includegraphics[width=0.9\linewidth]{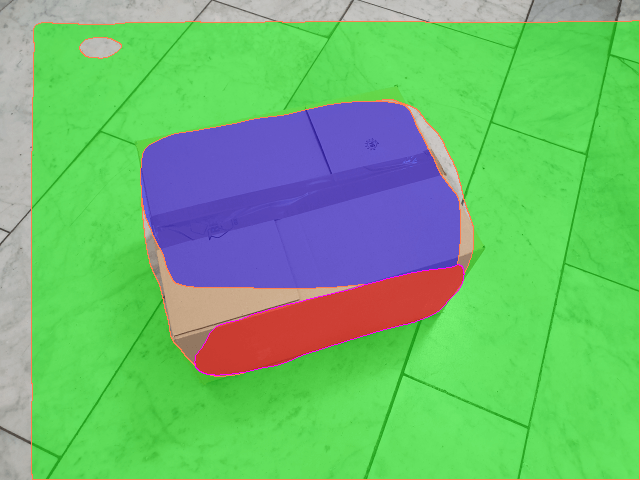} 
    \caption{Plane segmentation}
	\label{fig:overview:c}
  \end{subfigure}
  \begin{subfigure}[b]{0.5\linewidth}
    \centering
    \includegraphics[width=0.9\linewidth]{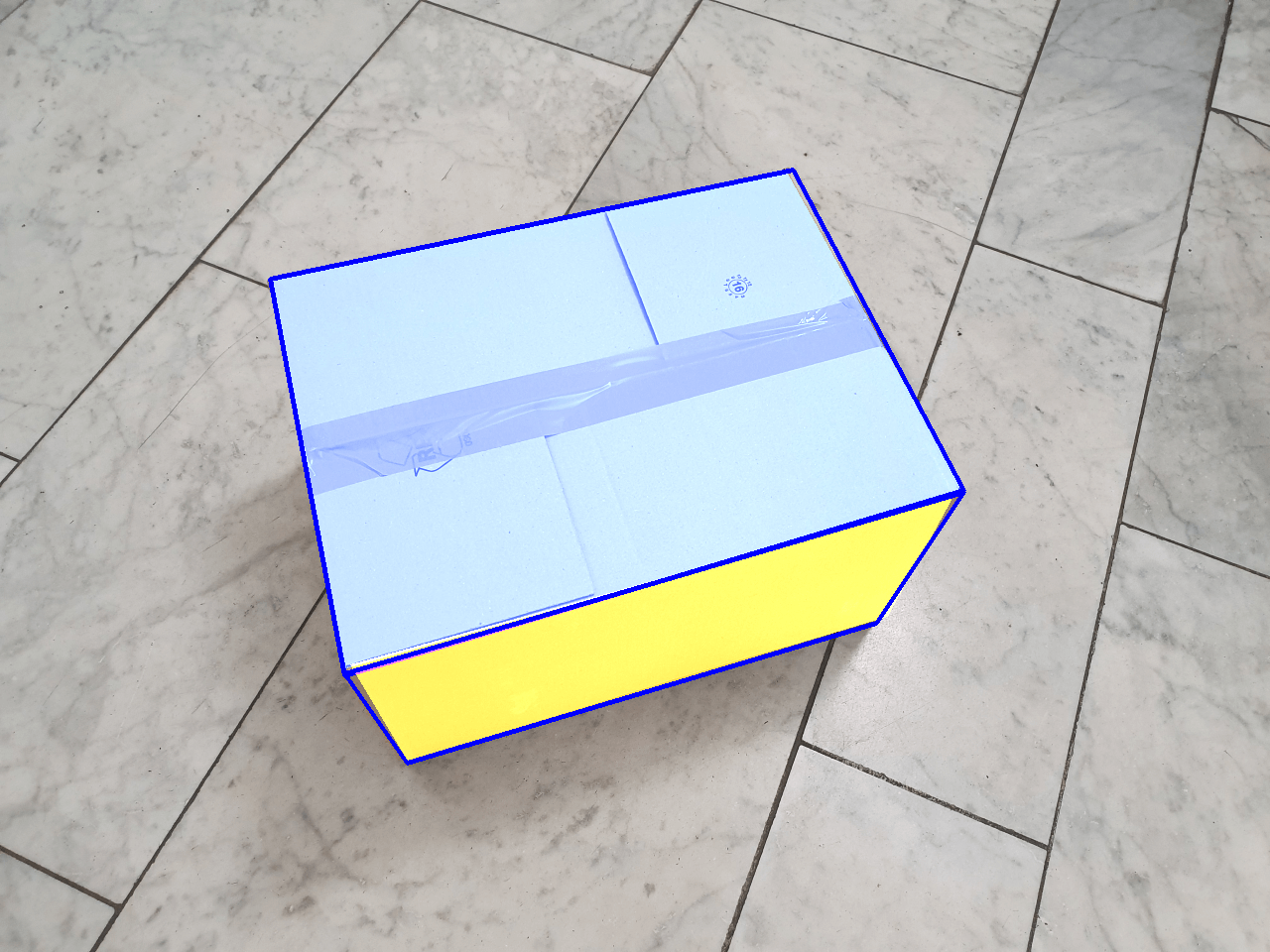} 
    \caption{Refined segmentation} 
	\label{fig:overview:d}
  \end{subfigure} 
  \caption{Overview of the segmentation pipeline. We consider an image (a) and combine edge detection (b) and plane segmentation (c) to obtain fine-grained plane segmentation results (d) close to the ground truth (blue contours).}
  \label{fig:overview} 
\end{figure}

Identifying planar regions is an important task, that can be used in the context of segmentation and reconstruction for 3D scenes.
Recent developments in the area of 3D plane detection from single RGB images \cite{liuPlaneNetPiecewisePlanar2018}, \cite{liuPlaneRCNN3DPlane2019} open up opportunities to employ such approaches for various applications.
\gls{ar} is one such application that can be used in numerous domains, ranging from logistics, manufacturing and military to education and entertainment \cite{yuUsefulVisualizationTechnique2010}.
Identified planes can be used to inpaint information and to place and simulate objects in a scene \cite{karschAutomaticSceneInference2014}.
For example in manufacturing, \gls{ar} can be used to simulate, assist and improve processes \cite{ongAugmentedRealityApplications2008}.
In addition to that, robotics is a broad area of application, where plane segmentation can help to navigate, grasp and perform various other tasks.
A common domain, where objects are confined to cuboid shapes is logistics.
One benefit of using plane segmentation for object detection in such environments is its robustness compared to \gls{sota} segmentation techniques \cite{heMaskRCNN2017} that rely on knowing instances of the object categories beforehand.
Thus, an accurate plane segmentation could be used for reconstruction or damage and tampering detection \cite{nocetiMulticameraSystemDamage2018} in logistics contexts.
Moreover, \gls{ar} can assist the packaging process to reduce error rates and document the process \cite{hochsteinPackassistentAssistenzsystemFuer2016}.
Using \gls{sota} plane segmentation techniques alone is suitable for many applications, such as partial inpainting, however, these techniques still have difficulties in delivering fine-grained segmentation results.
To overcome this issue, we propose a post-processing technique to refine the results of plane segmentation approaches such as the PlaneRCNN \cite{liuPlaneRCNN3DPlane2019}, to accurately detect the planes of cuboid-shaped objects in an image.
Plane segmentation masks are refined separately and thus, not only cuboid-shaped objects, such as packages, but any plane with a rhombic shape can be rectified.
In addition to that, the surfaces of those planes do not need to be completely flat, as in the case of a parcel, since also small 3D structures on the plane can be handled.
Our approach uses different edge segmentation techniques \cite{cannyComputationalApproachEdge1986}, \cite{soriaDenseExtremeInception2020} to align the masks resulting from the plane segmentation with the edges in the image.
For an overview of the pipeline, see \fig{fig:overview}.

We collected a dataset of 34 images in a logistics context with ground truth plane segmentations, which we made publicly available\footnote{Dataset available under \href{https://url.fzi.de/dataset_planeseg}{https://url.fzi.de/dataset_planeseg}.}.
The performance of our approach, compared to the PlaneRCNN baseline \cite{liuPlaneRCNN3DPlane2019} and a fallback routine which is also introduced in this work, is evaluated on this dataset.
We show an overall improvement of the averaged \gls{iou} of over $4.25$ percentage points compared to the baseline.
The quality of the refinement for individual images depends on the quality of the prior segmentation and the edge detection in the image.
Several examples are provided to illustrate the capabilities of the approach even in difficult settings, but also to point out areas for further improvements.
We will make our code publicly available under \href{https://url.fzi.de/refined_planeseg}{https://url.fzi.de/refined_planeseg}.

This work is structured as follows.
In \secr{sec:related_work} we will present an overview over related literature.
\secr{sec:approach} outlines our plane segmentation mask refinement approach and \secr{sec:results_and_evaluation} evaluates our approach by comparing it to two baselines on our newly collected dataset.
\secr{sec:conclusion} concludes the paper.
	\section{Related Work}
\label{sec:related_work}

To the best of our knowledge, there has been no approach yet to improve the performance of \gls{sota} plane segmentation techniques by leveraging additional information from edge detection techniques.

Edge detection is a field thoroughly studied in computer vision \cite{oskoeiSurveyEdgeDetection2010}.
\citeauthor{sobelCameraModelsMachine1972}'s work \cite{sobelCameraModelsMachine1972} was one of the early contributions, that inspired numerous edge detection techniques.
The Canny Edge Detector \cite{cannyComputationalApproachEdge1986} is one of those techniques, that was developed in \citeyear{cannyComputationalApproachEdge1986} and is still a very common choice to date.
Due to the popularity of the Canny Edge Detector, there has been a lot of research on improving it, for example by using an adaptive thresholds \cite{fangStudyApplicationOtsu2009}.
Recently, also \glspl{cnn} have been used to detect edges \cite{xieHolisticallyNestedEdgeDetection2015}, 
\cite{soriaDenseExtremeInception2020} in images.
These approaches often provide the advantage of being less sensitive to noise and are a promising alternative to classical techniques.

Image segmentation is a field intensively studied in computer vision that reached remarkable performance on closed-set configurations, where the objects of interests are known beforehand \cite{garcia-garciaReviewDeepLearning2017}.
In the area of plane segmentation, 
End-to-End trained \glspl{nn} have only been presented recently \cite{liuPlaneNetPiecewisePlanar2018}, \cite{liuPlaneRCNN3DPlane2019}.
In addition to improving on the \gls{sota} in plane segmentation, these approaches also improved the \gls{sota} in single-image depth estimation.

\section{Plane Segmentation Refinement}
\label{sec:approach}

Our approach aims to improve the granularity of the plane segmentation by exploiting additional information from edge detection.
We assume a prior segmentation by plane segmentation techniques, however, the procedure is independent of the source of these prior masks.
In the following, we introduce the edge detection techniques we consider in \secr{sec:approach:edge}.
Afterwards, we present the plane segmentation approach whose segmentation masks are the input for our post-processing in \secr{sec:approach:plane}.
Finally, we propose a method that leverages clustering and regression to find line segments along a rhombus in \secr{sec:approach:refine} and combine these ideas for the final refinement in \secr{sec:approach:approach}.
%

\subsection{Edge Detection}
\label{sec:approach:edge}
We consider two different techniques for edge segmentation.
A recent work building upon the Canny Edge Detector \cite{cannyComputationalApproachEdge1986} by automating the thresholding process \cite{fangStudyApplicationOtsu2009}, which we call Adaptive Canny and a machine learning based approach called DexiNed by \citeauthor{soriaDenseExtremeInception2020} \cite{soriaDenseExtremeInception2020}.
We use two different forms of the latter approach, by applying it to the full resolution image (1280x960) and to a downsized image (640x480), since this seems to trigger a focus on important edges.
%
%
%
%
%

\subsection{Plane Segmentation}
\label{sec:approach:plane}
The PlaneRCNN deep neural architecture was introduced by \citeauthor{liuPlaneRCNN3DPlane2019} in \citeyear{liuPlaneRCNN3DPlane2019} \cite{liuPlaneRCNN3DPlane2019}.
It improves upon earlier models \cite{liuPlaneNetPiecewisePlanar2018} \cite{yangRecovering3DPlanes2018} by not requiring the maximum number of planes a priori and generalizing better to unseen scenes.
The input to the model is a single RGB image, which is processed by three components.
The first component is a Mask R-CNN \cite{heMaskRCNN2017} based model for plane detection.
In addition to the plane segmentation, also the plane normal and depth values for each pixel are estimated.
The second and third component are responsible for a refinement and the enforcement of consistency of the reconstructions.
In this work, we use only the plane segmentations retrieved by the PlaneRCNN.

\subsection{Line Segmentation}
\label{sec:approach:refine}
\begin{figure}[ht!] 
  \begin{subfigure}[b]{0.5\linewidth}
    \centering
    \includegraphics[width=0.9\linewidth]{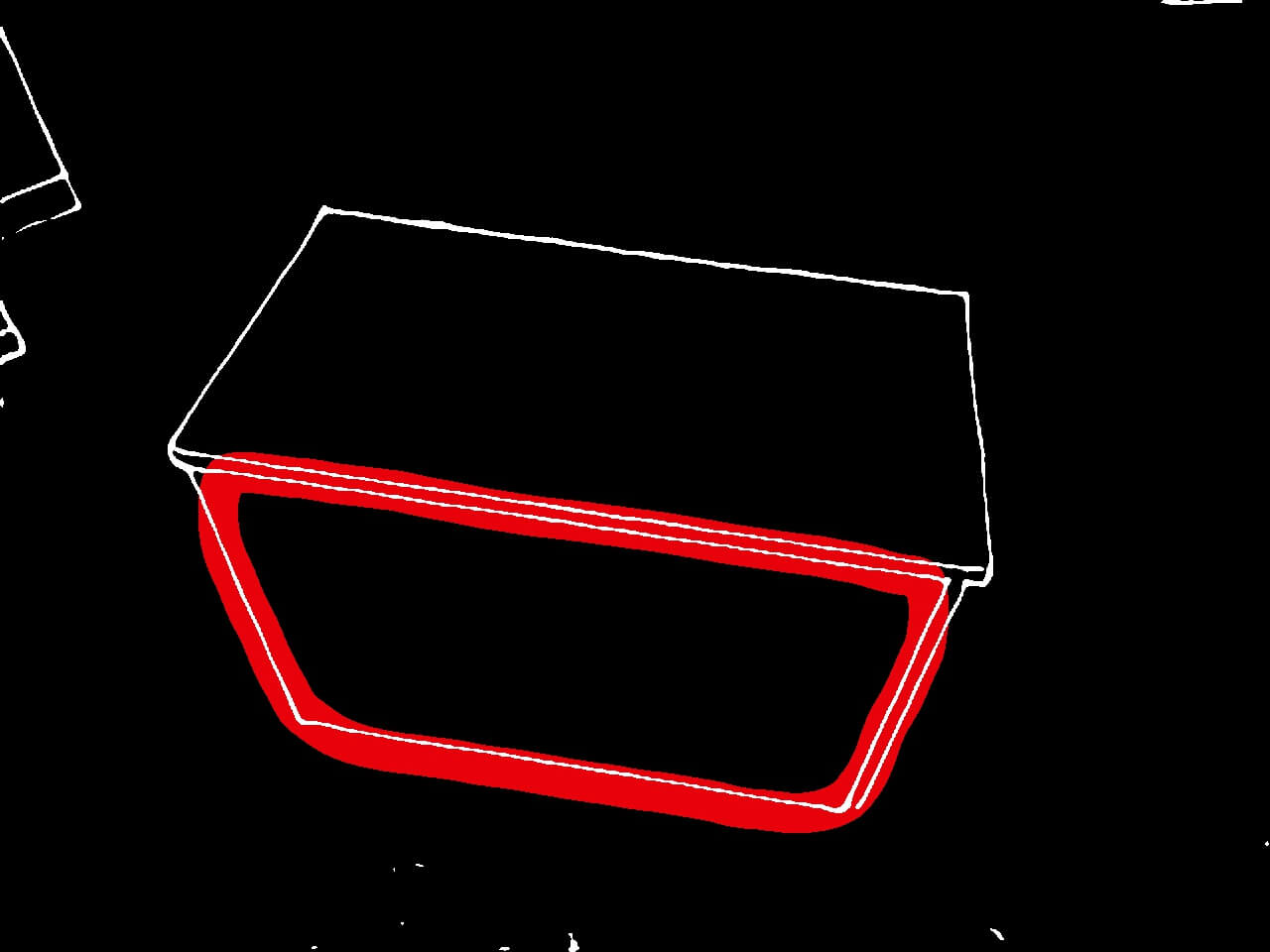} 
    \caption{Boundary of input mask} 
	\label{fig:line_detection:a}
    \vspace{2ex}
  \end{subfigure}
  \begin{subfigure}[b]{0.5\linewidth}
    \centering
    \includegraphics[width=0.9\linewidth]{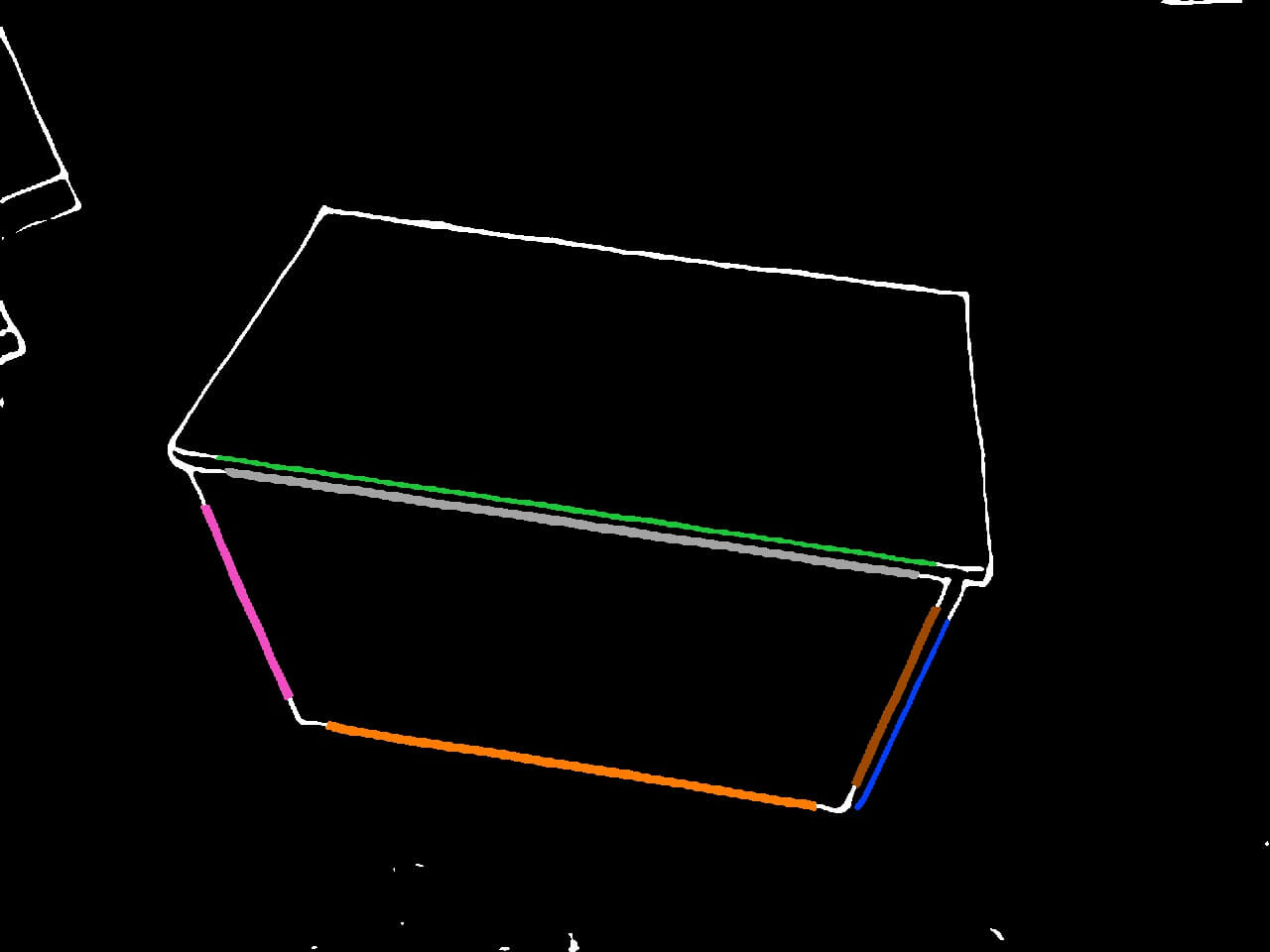} 
    \caption{Relevant segments after split} 
	\label{fig:line_detection:b}
    \vspace{2ex}
  \end{subfigure} 
  \begin{subfigure}[b]{0.5\linewidth}
    \centering
    \includegraphics[width=0.9\linewidth]{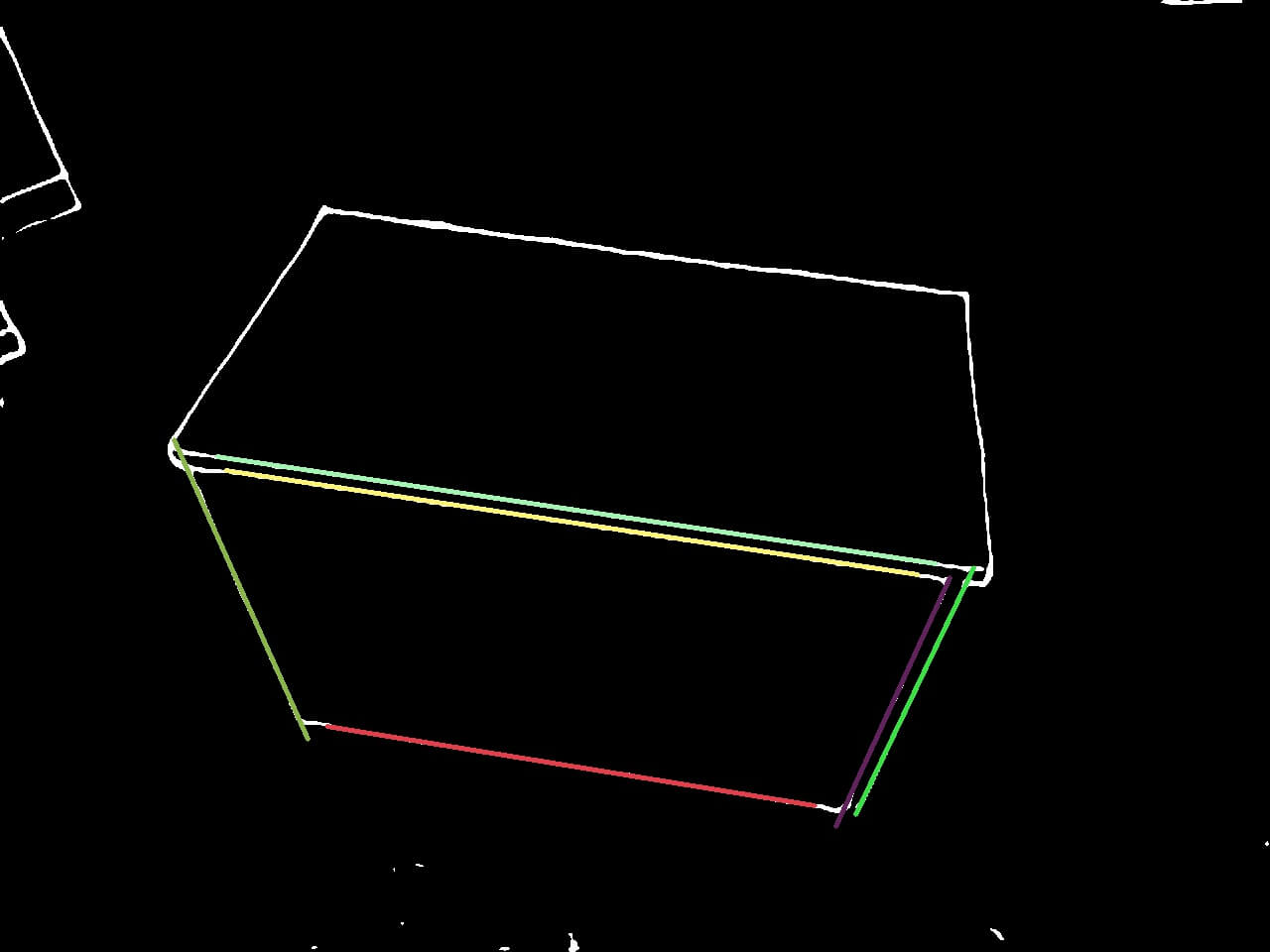} 
    \caption{Fitted lines}
	\label{fig:line_detection:c}
  \end{subfigure}
  \begin{subfigure}[b]{0.5\linewidth}
    \centering
    \includegraphics[width=0.9\linewidth]{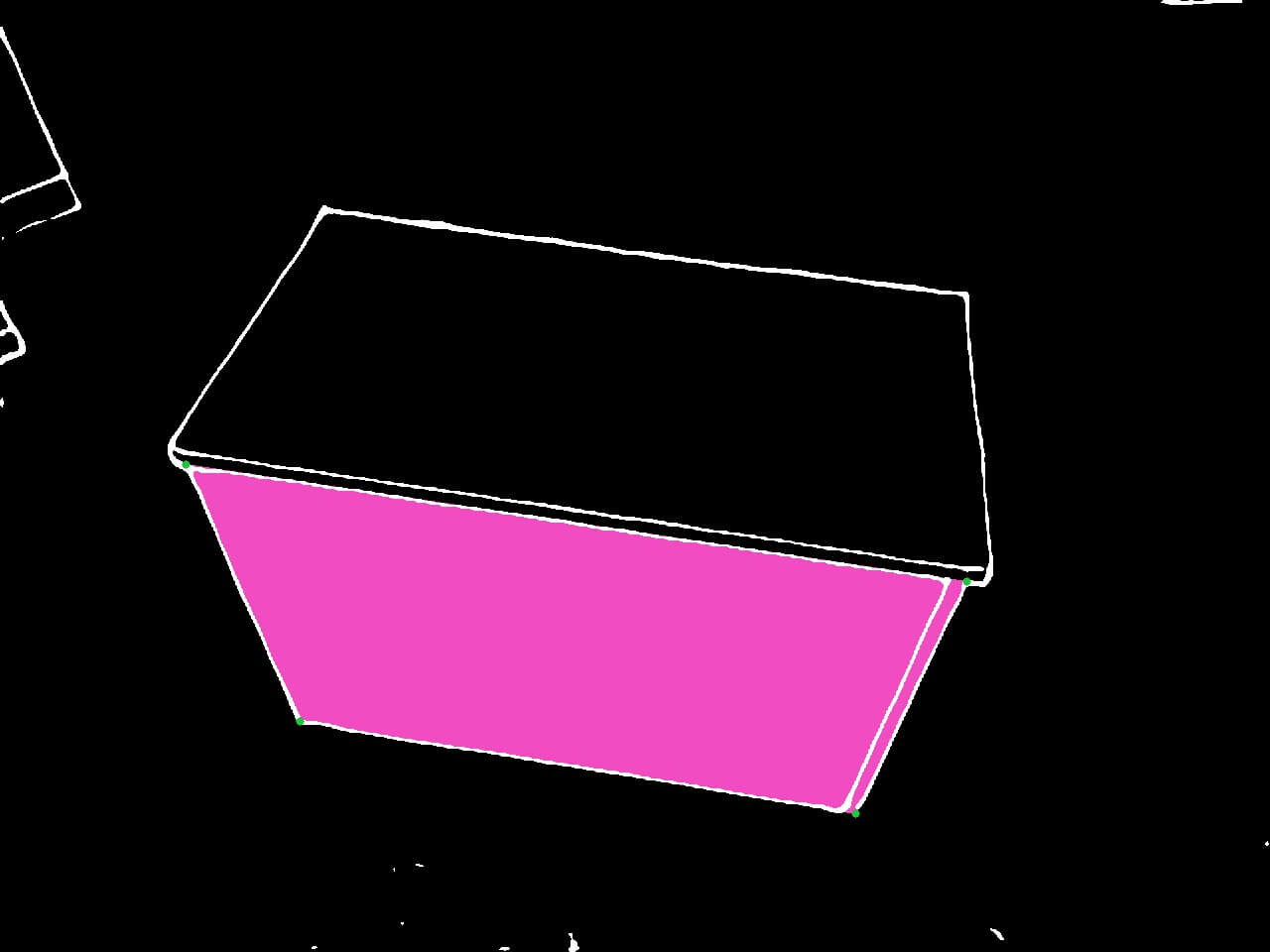} 
    \caption{Endpoints of lines} 
	\label{fig:line_detection:d}
  \end{subfigure} 
  \caption{Overview of the pipeline for the segmentation refinement. We use the widened contour of the mask (a) to identify clusters of line segments (b). We try to fit a line for each segment (c), which can be used for the estimation of the final segmentation mask (d).}
  \label{fig:line_detection} 
\end{figure}

Given a binary contour image and a segmentation of this image in planes, we consider the process of refining the given segmentation for each mask separately.
We present two approaches for detecting line segments on the bounding edges of a plane belonging to a cuboid-shaped object, which will be combined in the final solution.
The starting point for both approaches is to overlay the binary image with a widened contour line of the mask (See \fig{fig:line_detection:a}).
For a reasonable prior segmentation, this contour should contain all or at least some of the bounding edges for the respective plane.

The first approach relies on \gls{dbscan} \cite{esterDensitybasedAlgorithmDiscovering1996} to identify connected structures on the extract of the binary image.
For a good prior segmentation, this extract of the binary image might comprise a completely connected cluster or similarly, only two or three clusters for rhombus as seen in \fig{fig:line_detection:a}.
Since we try to estimate all line segments separately, we perform a corner detection \cite{harrisCombinedCornerEdge1988} and break the clusters up by removing the detected corners from the binary mask.
Thereafter, we are left with line-shaped clusters only as in \fig{fig:line_detection:b}, if all corners are detected correctly.
In addition to that, we omit very small clusters and clusters with big variance in two directions, since they most likely constitute areas of noise.
This leaves us with mainly line-shaped clusters of pixels.
We use a RANSAC \cite{fischlerRandomSampleConsensus1987} linear regression to find a two point description for each of those lines.
For each line, this first estimation is used to search for extensions of the line that where not captured by the widened contour line of the mask.
This becomes necessary when the widened contour line only overlays with parts of the current edge, since non-overlapping contours were previously ignored.
Hence, we create a mask along the estimated line across the whole image and repeat the former process.
We apply clustering onto the new mask and perform a RANSAC linear regression on the dominant cluster to obtain new end point estimations for the current line.
\scomment{Finally, to get the starting and end point of the line we use the minimum and maximum $x$ values.}

The second approach is based on the Hough Transform \cite{houghMethodMeansRecognizing1962}.
We cluster and average the lines resulting from applying the Hough Transform to the extract of the binary image.
This approach is complementary to the first approach, since it does not rely on the connectedness within edge segments.
Using the resulting lines in normal form, we estimate start and end points by leveraging their points of intersection.
More precisely, for each line we compute its points of intersection with all other lines and assume that valid endpoints lie in the vicinity of the considered mask.
The check for vicinity is performed by examining if a circle around the point of intersection with a 40 pixel radius intersects with the extract of the binary image.

\subsection{Combined Approach}
\label{sec:approach:approach}
Still considering each plane separately, we combine the results from the first and the second approach from \secr{sec:approach:refine}.
The result from each approach is a list of line segments described by a start and an end point, which represent approximations of the edges.
We cluster the line segments\scomment{in normal form} to identify lines describing the same edge.

By using a k-means algorithm, we group start points into a set $A$ and end points into a set $B$.
These groups are each complemented by a set of ten randomly chosen points on the binary image in the vicinity of the centroids resulting from the k-means algorithm.
We then identify the start point $P_a \in A$ and the end point $P_b \in B$ best fitting the considered edge $e$, for each cluster of line segments.
Our cost function $C$ describing the quality of the fit incorporates the overlap with the underlying mask $m_e$ and the normalized length of the line with equal weights
\[
C(P_a, P_b) = 0.5 \cdot \text{I}(\overline{P_aP_b}, m_e) + 0.5 \cdot \frac{\norm{P_a - P_b}}{\text{max}_{k \in A, l \in B}(\norm{P_k - P_l})},
\]
where $\text{I}(\overline{P_aP_b}, m_e)$ is the normalized intersection
\[
\text{I}(\overline{P_aP_b}, m_e) = \frac{\overline{P_aP_b} \cap m_e}{\overline{P_aP_b}}.
\]
The area $\overline{P_aP_b}$ is defined by a line with one pixel thickness between $P_a$ and $P_b$. 
Note that focusing on maximum overlap only might lead to very accurate, however, contracted line segments that do not represent the edges of the plane well.

Since we are aware of the dependence of our approach on the quality of the input information, we check the consistency of the refined mask with the prior mask by calculating their \gls{iou}.
For large deviations, i.e. an \gls{iou} of less than $0.75$, we resort back to a simple baseline approach.
This baseline consist of calculating the prior mask's convex hull and iteratively reducing the set of points describing the mask \cite{douglasAlgorithmsReductionNumber1973} to 20 points or less.

Our approach is based on considering one mask at a time. Note that the dependencies between masks belonging to the same object can be used to further refine the segmentation results.

	\section{Results and Evaluation}
\label{sec:results_and_evaluation}

We first describe the dataset we collected in \secr{sec:eval:dataset} and comment on its separation into different classes.
Subsequently, we will shortly present the baseline approaches used in the evaluation and finally discuss the results of our approach.

\subsection{Dataset}
\label{sec:eval:dataset}

We collected a set of 34 images, each picturing one cuboid object with different backgrounds.
The backgrounds include carpet, table surfaces, floor tiles and the inside of a container.
The cuboid-shaped objects include different types of parcels made from carton and plastic.
The surfaces of the parcels range from almost plain to complex structures.

During the process of collecting the images, we manually discarded all images where the PlaneRCNN was not able to grasp the scene, i.e. where it did not detect all visible planes of the cuboid object of interest.
This was mostly the case for flat objects and difficult camera angles, however, it also happened in some simpler scenes.
We split up the dataset in three groups by manually assessing the complexity of the scenes, i.e. the quality of the prior segmentation masks and the edge detection.
Of the 34 images, 7 were grouped into the category easy, 16 were grouped into the category medium and the remaining 11 were assigned the difficulty hard.
Note that even the easy category contains diverse backgrounds and different types of packages.
The data was labeled manually using the VGG Image Annotator \cite{duttaAnnotationSoftwareImages2019}.

\subsection{Baseline Approaches}
\label{sec:eval:baseline}

The results from the PlaneRCNN \cite{liuPlaneRCNN3DPlane2019} are used as input for our approach and thus, constitute a first baseline.
In addition to that, we present the results of using our fallback solution that was described in \secr{sec:approach:approach}.
Results separated by categories are presented in Table \ref{table:results}.
We use the \gls{iou} as metric for comparison as common for segmentation tasks.
The \gls{iou} over all masks in an image is averaged and subsequently the average over all images in the dataset is taken.

As mentioned above, we removed images from the dataset, where the PlaneRCNN was not able to grasp the scene.
If the PlaneRCNN is able to grasp the scene, it reaches 81.94\% \gls{iou} with the ground truth masks.
Our fallback solution shows a slight improvement on the PlaneRCNN by $0.23$ percentage points for the \gls{iou}.
Since it rectifies the PlaneRCNN masks, their shape is more consistent with the ground truth masks.

\begin{table}
	\centering
	\begin{tabular}{|c|c|c|c|c|c|}
		\hline 
		Dataset & PlaneRCNN & Fallback & Dexi LR & Dexi FR  & Canny\\ 
		\hline 
		Easy & 83.85\% & 84.00\% & \textbf{90.74\%} & 87.16\% & 82.03\% \\ 
		Medium & 82.52\% & 82.67\% & \textbf{84.75\%} & 82.66\% & 81.91\% \\ 
		Hard & 79.88\% & 80.29\% & \textbf{82.38\%} & 80.43\% & 79.42\% \\ 
		\hline 
		All & 81.94\% & 82.17\% & \textbf{85.20\%} & 83.10\% & 81.07\% \\
		\hline 
	\end{tabular}
	\caption{Average \gls{iou} over all masks in each dataset for the PlaneRCNN, the fallback solution and our suggested approach with different edge segmentation techniques (Dexi LR = DexiNed with low resolution, Dexi FR = DexiNed with full resolution, Canny = Adaptive Canny).}
	\label{table:results}
\end{table}

\subsection{Our Approach}
\label{sec:eval:our}

\begin{figure}[ht!] 
  \begin{subfigure}[b]{0.5\linewidth}
    \centering
    \includegraphics[width=0.45\linewidth]{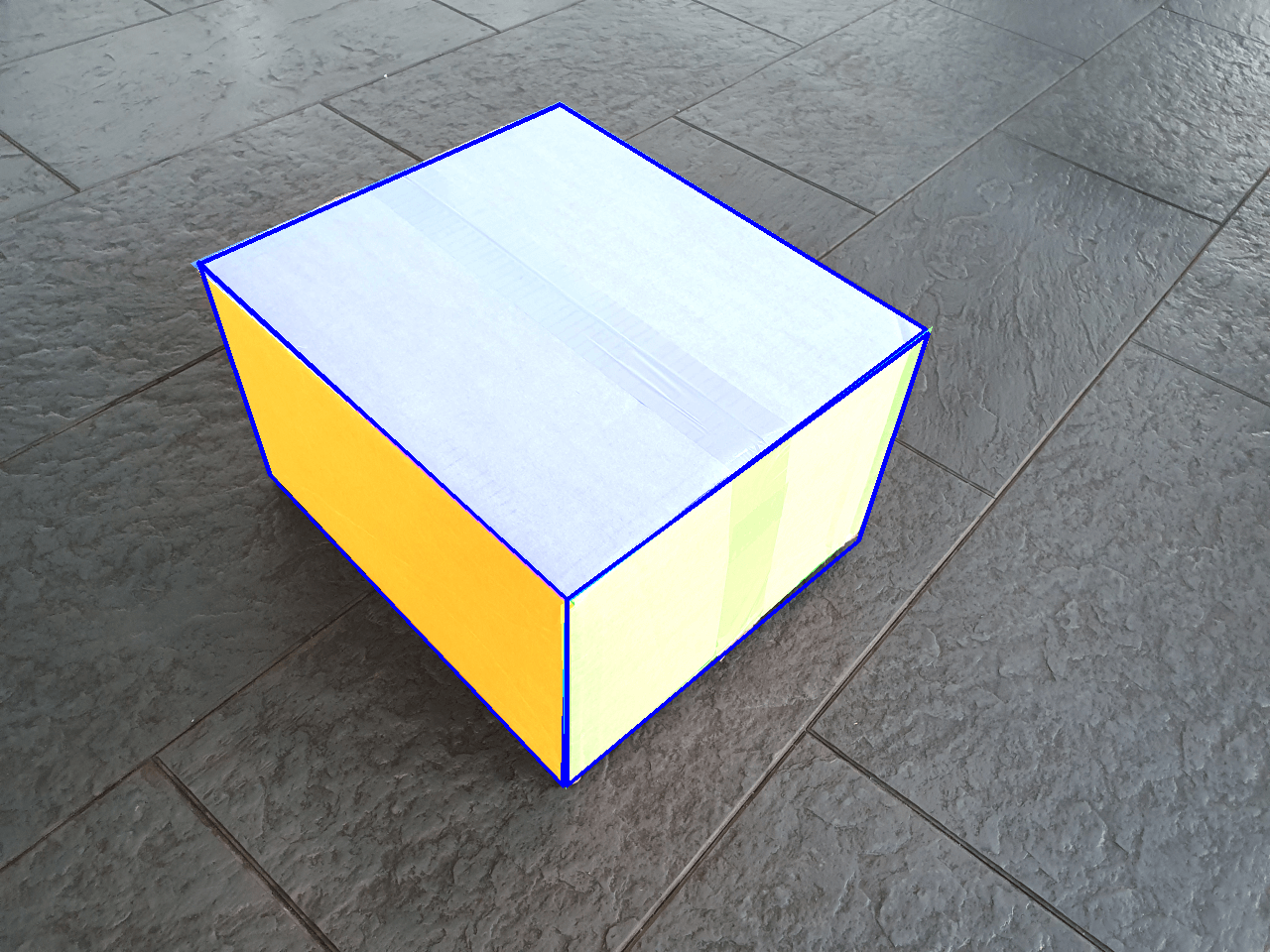} 
    \includegraphics[width=0.45\linewidth]{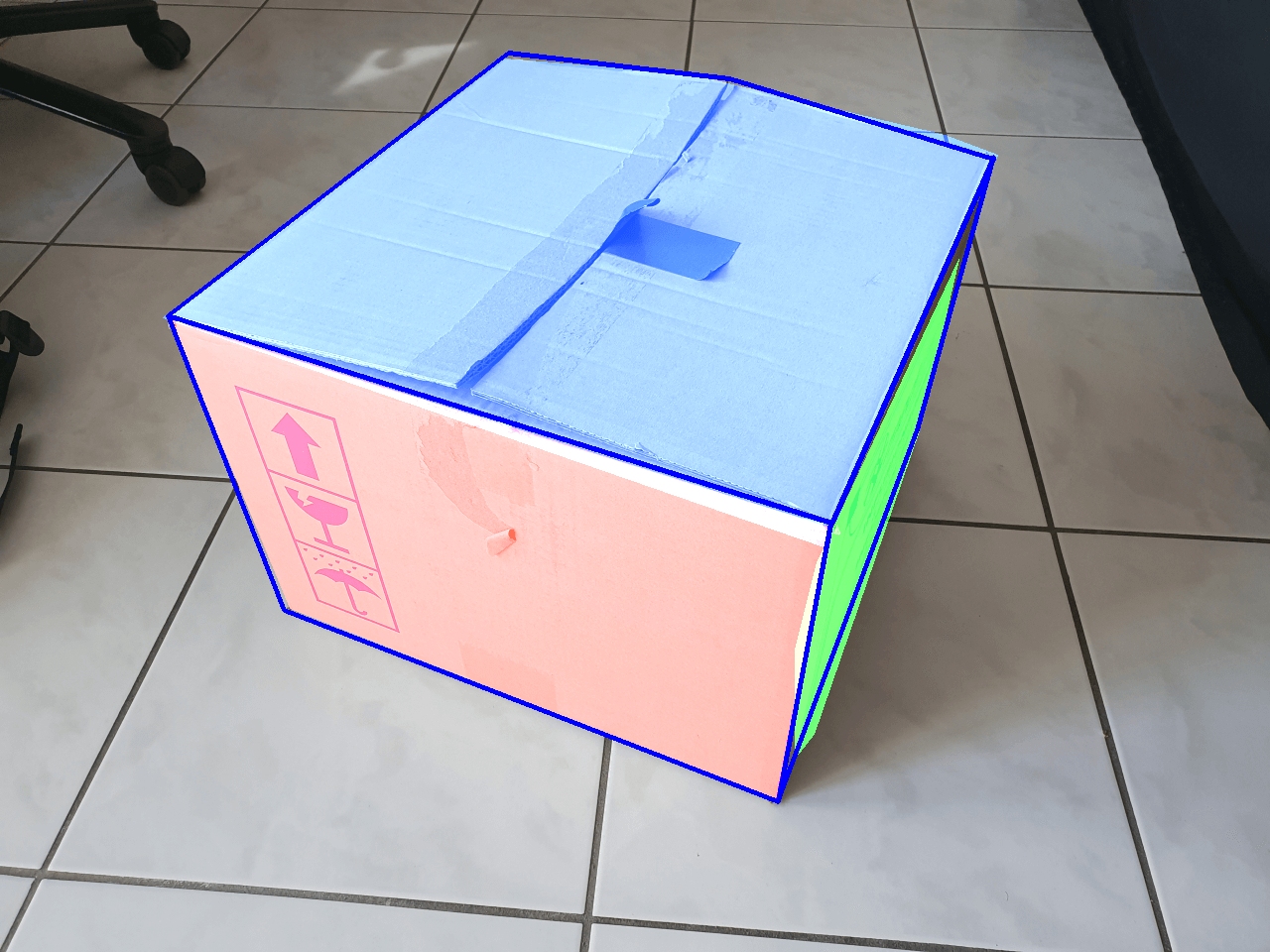} 
    \caption{Classified as easy.} 
	\label{fig:good:a}
  \end{subfigure}
  \begin{subfigure}[b]{0.5\linewidth}
    \centering
    \includegraphics[width=0.45\linewidth]{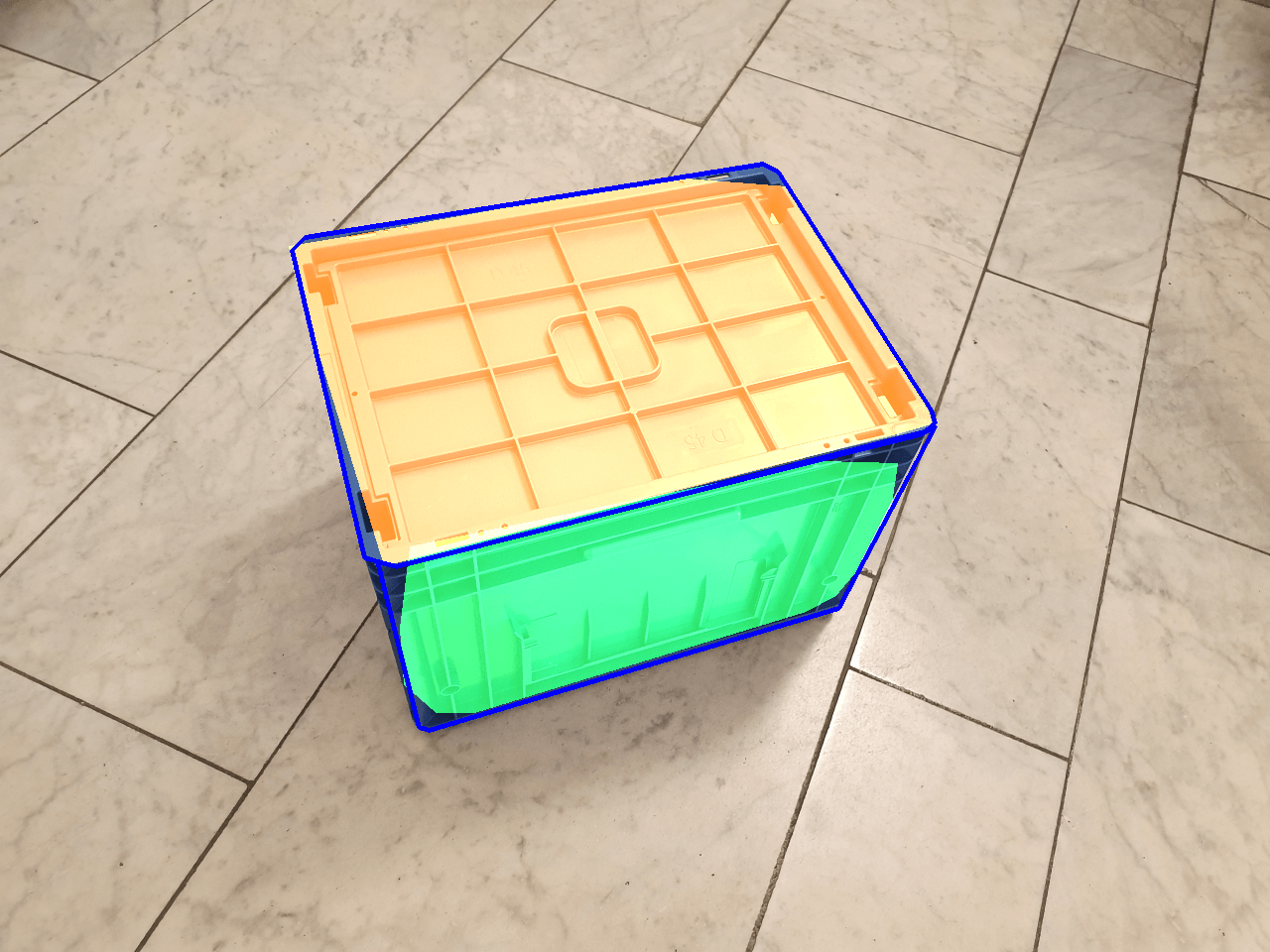} 
    \includegraphics[width=0.45\linewidth]{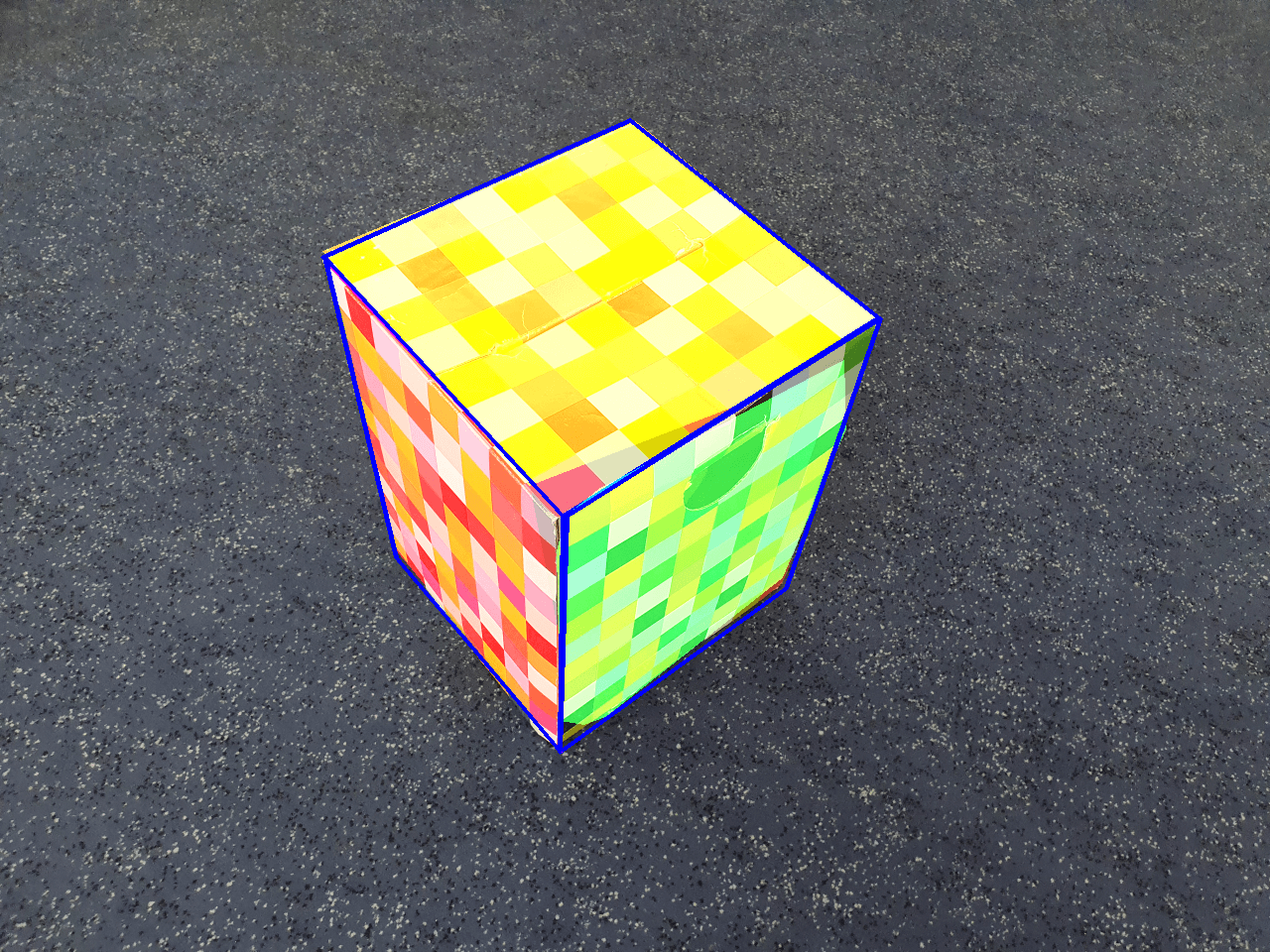} 
    \caption{Classified as hard.} 
	\label{fig:good:b}
  \end{subfigure} 
  \caption{Exemplary results of the plane segmentation refinement. The blue contours constitute the ground truth.}
  \label{fig:good} 
\end{figure}

The evaluation results for our approach using different edge detection techniques are reported in Table \ref{table:results}.
The evaluation shows an improvement of over $4.25$ percentage points compared to the PlaneRCNN when edge detection is performed by DexiNed with a low resolution image as input.
The results on the dataset classified as easy show an improvement of almost $7$ percentage points.
Thus, especially for reasonable prior segmentation masks and edge detections a considerable improvement over the baseline can be achieved.
We exemplarily show such segmentation results in \fig{fig:overview} and \fig{fig:good}.
Note that, even for complex backgrounds and feature-rich objects, our approach achieves good accuracy.

The importance of the edge detection technique is seen by its strong influence on the evaluation results.
We observe a consistent decline over all datasets moving from DexiNed with low resolution to DexiNed with full resolution and from DexiNed with full resolution to the Adaptive Canny algorithm.
Using the adaptive Canny Edge Detector even yields results slightly worse than the baseline, as can be seen in Table \ref{table:results}.
This is due to feature-rich backgrounds and parcels where the bounding edges are not dominant.

\begin{figure}[ht!] 
  \begin{subfigure}[b]{0.24\linewidth}
    \centering
    \includegraphics[width=0.95\linewidth]{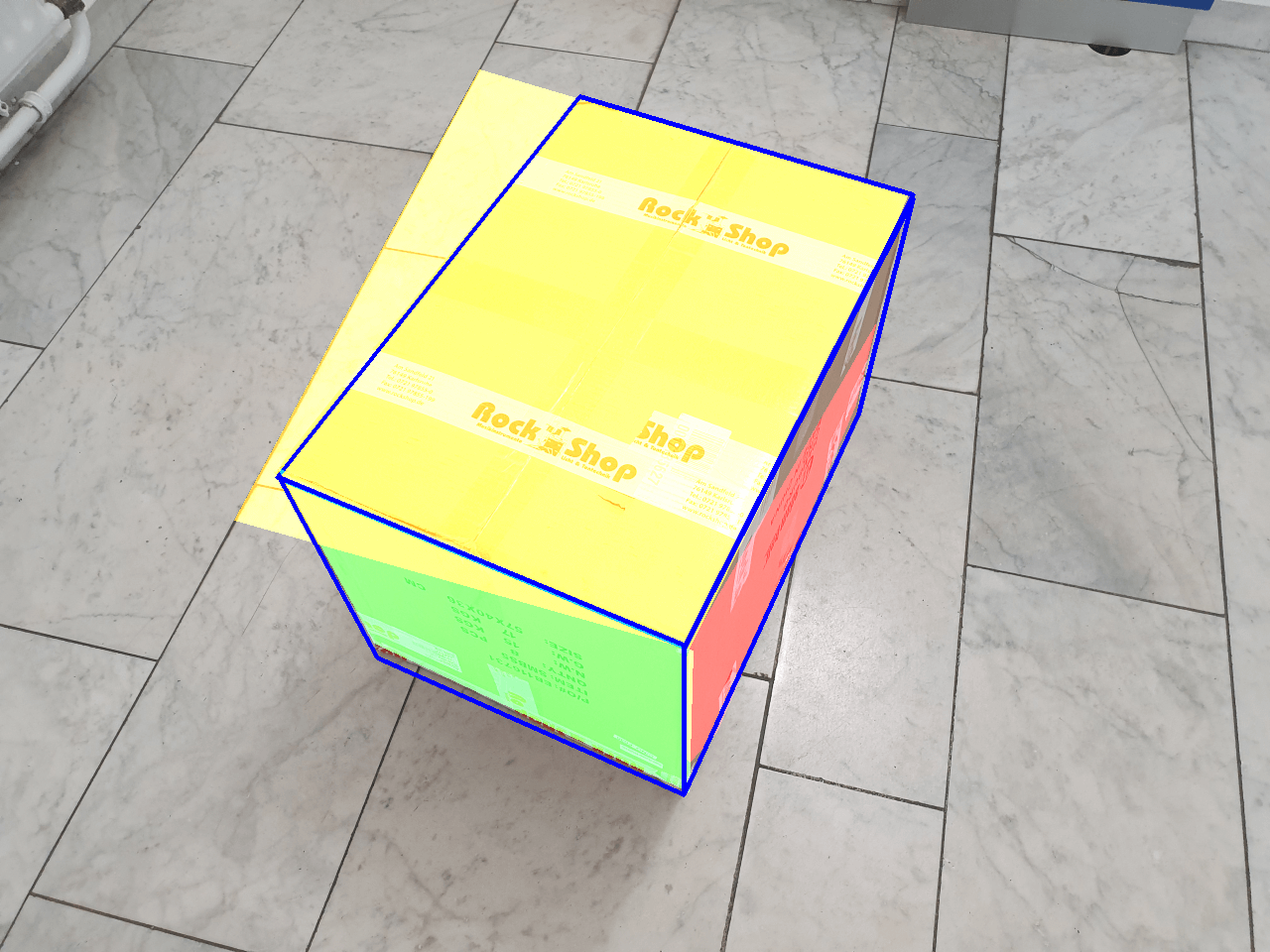} 
    \caption{}
	\label{fig:issues:a}
  \end{subfigure}
  \begin{subfigure}[b]{0.24\linewidth}
    \centering
    \includegraphics[width=0.95\linewidth]{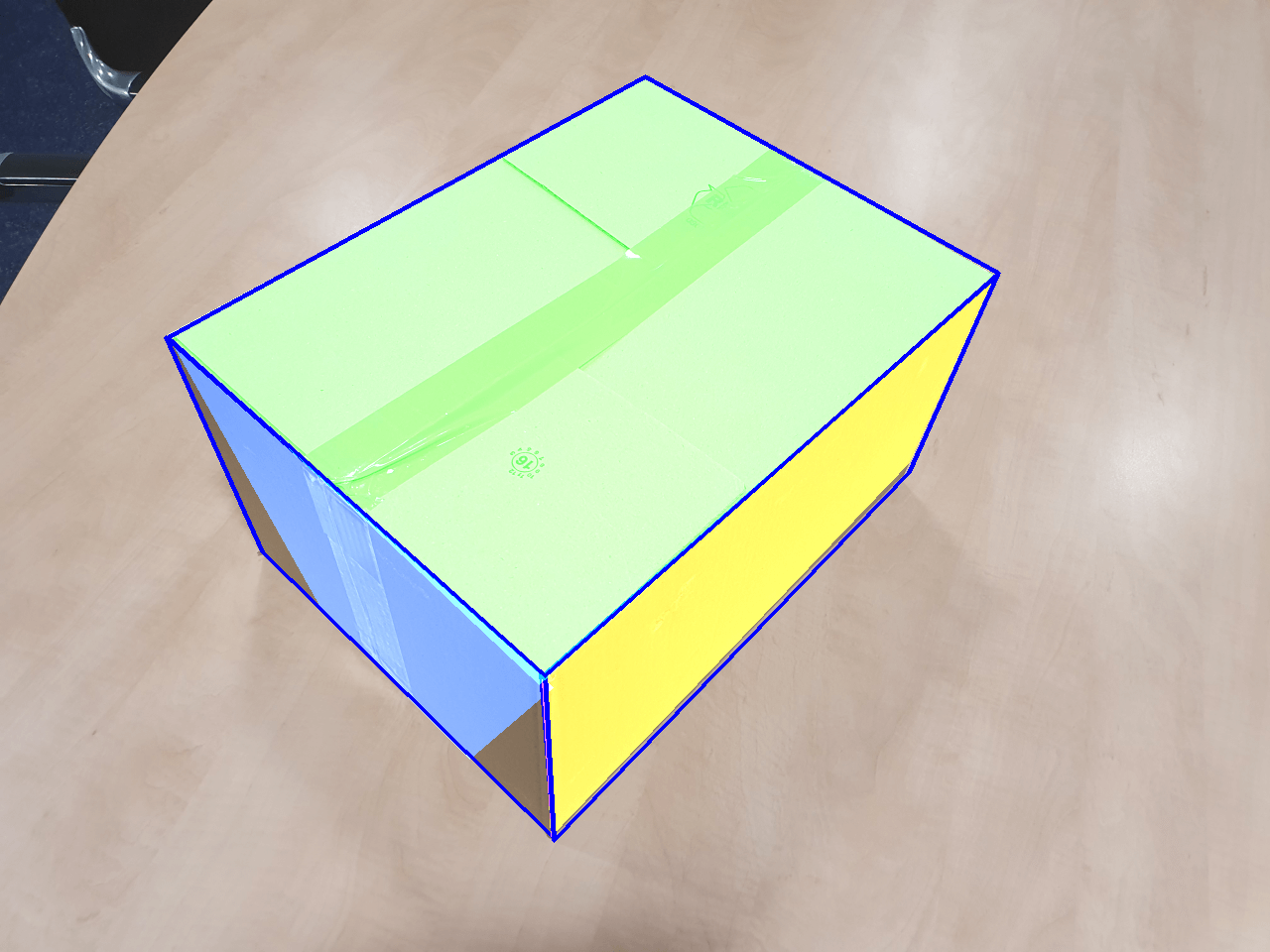} 
    \caption{}
	\label{fig:issues:b}
  \end{subfigure}
  \begin{subfigure}[b]{0.24\linewidth}
    \centering
    \includegraphics[width=0.95\linewidth]{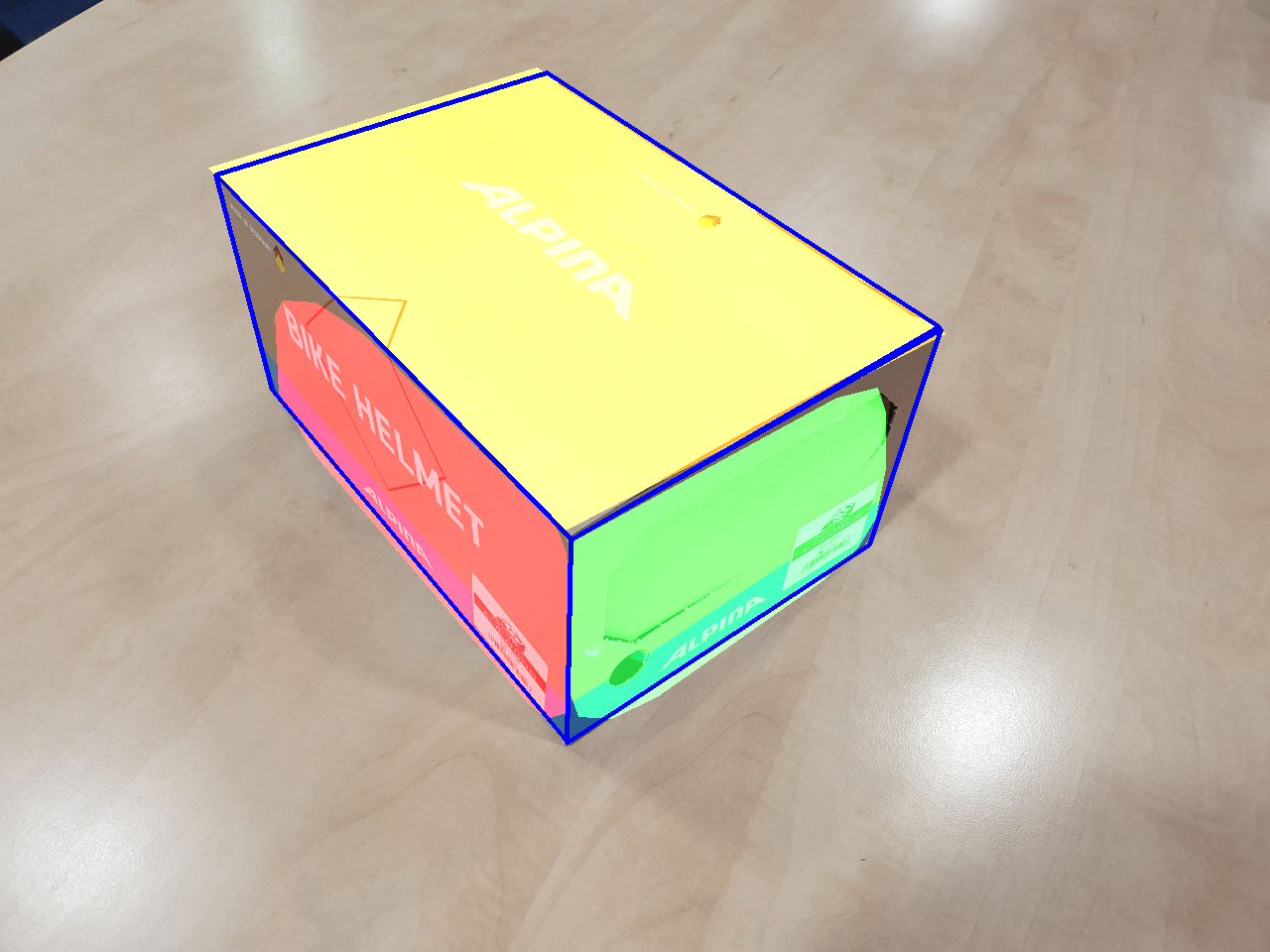} 
    \caption{}
	\label{fig:issues:c}
  \end{subfigure}
  \begin{subfigure}[b]{0.24\linewidth}
    \centering
    \includegraphics[width=0.95\linewidth]{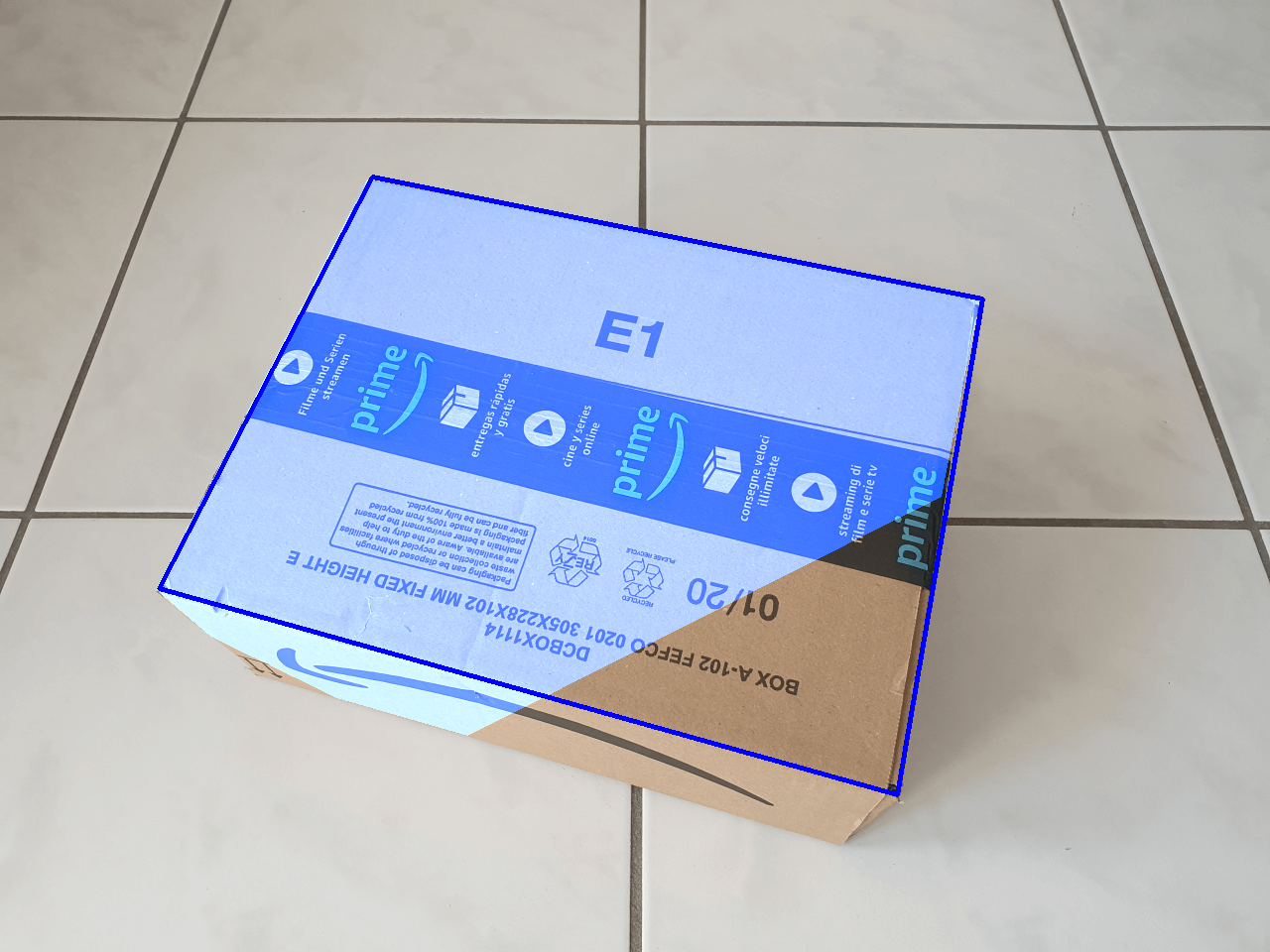} 
    \caption{}
	\label{fig:issues:d}
  \end{subfigure} 
  \caption{Visualization of the dependence of our approach on accurate prior information. 
  Examples with strong background edges (a), failed corner detection (b), too small prior segmentation (c) and too big prior segmentation (d).}
  \label{fig:issues} 
\end{figure}

By manually assessing the images, we identified reasons for bad segmentation results.
The approach performs well when the prior mask and the edge detection yield reasonable inputs.
Strong edges in the background as in \fig{fig:issues:a} and dominant lines on the parcel as the black label in \fig{fig:issues:d} can sometimes mislead the algorithm and cause deviations from the desired mask.
In addition to that, the ability to break up the clustered segments around the mask by detecting and removing corners does affect the accuracy.
In \fig{fig:issues:b}, for example, the vertical lines of the blue plane were not detected since they were clustered with the more dominant horizontal lines.
Finally, imprecise prior masks can cause wrongfully merging two planes of an object as in \fig{fig:issues:d} or prohibit the algorithm from detecting the full surface of the plane as in \fig{fig:issues:c}.

\section{Conclusion}
\label{sec:conclusion}

In this work, we presented an approach for the refinement of segmentation masks of cuboid-shaped objects.
Existing \gls{sota} plane segmentation methods generalize well for the logistics environment we considered exemplarily and often yielded reasonable scene segmentations.
However, those segmentation masks lack accuracy for fine-grained details such as corners.
To enable the use of plane segmentation techniques for a wider range of applications, where this accuracy is necessary, we propose a post-processing technique.
We combine edge detection and plane segmentation techniques with clustering approaches to perform a mask refinement along the edges of the object.
To achieve robustness, we complement these techniques with a simple, yet effective fallback solution. 
Our approach improves the accuracy of \gls{sota} plane segmentation techniques by over $4.25$ percentage points and generates masks that are more consistent in shape with the ground truth masks.
The refined segmentation masks have several applications, for instance inpainting for \gls{ar} or object detection for cuboid-shaped objects.
The further improvement of the segmentation by exploiting the fact that planes belong to the same object is left for future research.

\printbibliography

@article{douglasAlgorithmsReductionNumber1973,
  title = {Algorithms for the Reduction of the Number of Points Required to Represent a Digitized Line or Its Caricature},
  author = {Douglas, David H and Peucker, Thomas K},
  date = {1973},
  journaltitle = {Cartographica: The International Journal for Geographic Information and Geovisualization},
  shortjournal = {Cartographica: The International Journal for Geographic Information and Geovisualization},
  volume = {10},
  pages = {112--122},
  doi = {10/bjwv52},
  url = {https://utpjournals.press/doi/10.3138/FM57-6770-U75U-7727},
  urldate = {2020-03-02},
  langid = {english},
  number = {2}
}

@article{cannyComputationalApproachEdge1986,
  title = {A {{Computational Approach}} to {{Edge Detection}}},
  author = {Canny, John},
  date = {1986},
  journaltitle = {IEEE Transactions on Pattern Analysis and Machine Intelligence},
  pages = {679--698},
  issn = {1939-3539},
  doi = {10/fn3fdk},
  number = {6}
}

@inproceedings{duttaAnnotationSoftwareImages2019,
  title = {The {{VIA Annotation Software}} for {{Images}}, {{Audio}} and {{Video}}},
  booktitle = {{{ACM International Conference}} on {{Multimedia}}},
  author = {Dutta, Abhishek and Zisserman, Andrew},
  date = {2019},
  pages = {2276--2279},
  publisher = {{ACM}},
  doi = {10/ggk524},
  url = {http://dl.acm.org/doi/10.1145/3343031.3350535},
  urldate = {2020-02-19},
  eventtitle = {{{MM}} '19: {{The}} 27th {{ACM International Conference}} on {{Multimedia}}},
  langid = {english},
}

@inproceedings{esterDensitybasedAlgorithmDiscovering1996,
  title = {A Density-Based Algorithm for Discovering Clusters a Density-Based Algorithm for Discovering Clusters in Large Spatial Databases with Noise},
  booktitle = {Proceedings of the {{Second International Conference}} on {{Knowledge Discovery}} and {{Data Mining}}},
  author = {Ester, Martin and Kriegel, Hans-Peter and Sander, Jörg and Xu, Xiaowei},
  date = {1996},
  pages = {226--231},
  location = {{Portland, Oregon}},
  series = {{{KDD}}'96}
}

@article{fangStudyApplicationOtsu2009,
  title = {The {{Study}} on {{An Application}} of {{Otsu Method}} in {{Canny Operator}}},
  author = {Fang, Mei and Yue, GuangXue and Yu, QingCang},
  date = {2009},
  journaltitle = {International Symposium on Information Processing},
  pages = {4},
  langid = {english},
  series = {Proceedings / the 2009 {{International Symposium}} on {{Information Processing}}: 21 - 23 {{August}} 2009, {{Huangshan}}, {{China}}}
}

@incollection{fischlerRandomSampleConsensus1987,
  title = {Random {{Sample Consensus}}: {{A Paradigm}} for {{Model Fitting}} with {{Applications}} to {{Image Analysis}} and {{Automated Cartography}}},
  shorttitle = {Random {{Sample Consensus}}},
  booktitle = {Readings in {{Computer Vision}}},
  author = {Fischler, Martin A. and Bolles, Robert C.},
  date = {1987},
  pages = {726--740},
  publisher = {{Elsevier}},
  doi = {10.1016/B978-0-08-051581-6.50070-2},
  url = {https://linkinghub.elsevier.com/retrieve/pii/B9780080515816500702},
  urldate = {2019-11-28},
  isbn = {978-0-08-051581-6},
  langid = {english},
}

@article{garcia-garciaReviewDeepLearning2017,
  title = {A {{Review}} on {{Deep Learning Techniques Applied}} to {{Semantic Segmentation}}},
  author = {Garcia-Garcia, Alberto and Orts-Escolano, Sergio and Oprea, Sergiu and Villena-Martinez, Victor and Garcia-Rodriguez, Jose},
  date = {2017},
  url = {http://arxiv.org/abs/1704.06857},
  urldate = {2019-03-06},
  archivePrefix = {arXiv},
  eprint = {1704.06857},
  eprinttype = {arxiv},
}

@inproceedings{harrisCombinedCornerEdge1988,
  title = {A {{Combined Corner}} and {{Edge Detector}}},
  booktitle = {Procedings of the {{Alvey Vision Conference}}},
  author = {Harris, C. and Stephens, M.},
  date = {1988},
  pages = {23.1-23.6},
  publisher = {{Alvey Vision Club}},
  location = {{Manchester}},
  doi = {10/gfvtg5},
  url = {http://www.bmva.org/bmvc/1988/avc-88-023.html},
  urldate = {2019-11-29},
  eventtitle = {Alvey {{Vision Conference}} 1988},
  langid = {english},
}

@inproceedings{heMaskRCNN2017,
  title = {Mask {{R}}-{{CNN}}},
  booktitle = {{{IEEE International Conference}} on {{Computer Vision}} },
  author = {He, Kaiming and Gkioxari, Georgia and Dollar, Piotr and Girshick, Ross},
  date = {2017},
  pages = {2980--2988},
  doi = {10.1109/ICCV.2017.322},
  url = {http://ieeexplore.ieee.org/document/8237584/},
  urldate = {2019-05-23},
  eventtitle = {2017 {{IEEE International Conference}} on {{Computer Vision}} },
  isbn = {978-1-5386-1032-9},
}

@inproceedings{hochsteinPackassistentAssistenzsystemFuer2016,
  title = {Packassistent – Assistenzsystem für die Qualitätskontrolle während des Packprozesses},
  booktitle= {Logistics Journal},
  author = {Hochstein, Maximilian and Glöckle, Johannes and Meyer, Thomas and Furmans, Kai},
  date = {2016},
  publisher = {{Wiss. Gesellschaft für Technische Logistik}},
  doi = {10.2195/lj_proc_hochstein_de_201610_01},
  url = {http://www.logistics-journal.de/proceedings/2016/fachkolloquium2016/4471},
  urldate = {2019-11-22},
  langid = {german},
}

@patent{houghMethodMeansRecognizing1962,
  title = {Method and Means for Recognizing Complex Patterns},
  author = {Hough, Paul V. C.},
  date = {1962},
  url = {https://patents.google.com/patent/US3069654A/en},
  urldate = {2020-02-07},
  holder = {{Paul V C Hough}},
  number = {3069654A},
  type = {patentus}
}

@article{karschAutomaticSceneInference2014,
  title = {Automatic {{Scene Inference}} for {{3D Object Compositing}}},
  author = {Karsch, Kevin and Sunkavalli, Kalyan and Hadap, Sunil and Carr, Nathan and Jin, Hailin and Fonte, Rafael and Sittig, Michael and Forsyth, David},
  date = {2014},
  journaltitle = {ACM Transactions on Graphics},
  shortjournal = {ACM Trans. Graph.},
  volume = {33},
  pages = {32:1--32:15},
  issn = {0730-0301},
  doi = {10/gd4vqm},
  url = {https://doi.org/10.1145/2602146},
  urldate = {2020-02-07},
  number = {3}
}

@inproceedings{liuPlaneNetPiecewisePlanar2018,
  title = {{{PlaneNet}}: {{Piece}}-{{Wise Planar Reconstruction}} from a {{Single RGB Image}}},
  shorttitle = {{{PlaneNet}}},
  booktitle = {{{IEEE Conference}} on {{Computer Vision}} and {{Pattern Recognition}}},
  author = {Liu, Chen and Yang, Jimei and Ceylan, Duygu and Yumer, Ersin and Furukawa, Yasutaka},
  date = {2018},
  pages = {2579--2588},
  issn = {1063-6919},
  doi = {10/ggm6gj},
  eventtitle = {2018 {{IEEE Conference}} on {{Computer Vision}} and {{Pattern Recognition}}},
}

@inproceedings{liuPlaneRCNN3DPlane2019,
  title = {{{PlaneRCNN}}: {{3D Plane Detection}} and {{Reconstruction From}} a {{Single Image}}},
  shorttitle = {{{PlaneRCNN}}},
  booktitle = {{{IEEE Conference}} on {{Computer Vision}} and {{Pattern Recognition}}},
  author = {Liu, Chen and Kim, Kihwan and Gu, Jinwei and Furukawa, Yasutaka and Kautz, Jan},
  date = {2019},
  pages = {4445--4454},
  location = {{Long Beach, CA, USA}},
  doi = {10/ggkg6t},
  url = {https://ieeexplore.ieee.org/document/8953257/},
  urldate = {2020-02-07},
  eventtitle = {2019 {{IEEE Conference}} on {{Computer Vision}} and {{Pattern Recognition}}},
  isbn = {978-1-72813-293-8},
  langid = {english},
}

@article{nocetiMulticameraSystemDamage2018,
  title = {A Multi-Camera System for Damage and Tampering Detection in a Postal Security Framework},
  author = {Noceti, Nicoletta and Zini, Luca and Odone, Francesca},
  date = {2018},
  journaltitle = {EURASIP Journal on Image and Video Processing},
  shortjournal = {EURASIP Journal on Image and Video Processing},
  volume = {2018},
  pages = {11},
  issn = {1687-5281},
  doi = {10.1186/s13640-017-0242-x},
  url = {https://doi.org/10.1186/s13640-017-0242-x},
  urldate = {2019-11-22},
  number = {1}
}

@article{ongAugmentedRealityApplications2008,
  title = {Augmented Reality Applications in Manufacturing: A Survey},
  shorttitle = {Augmented Reality Applications in Manufacturing},
  author = {Ong, S. K. and Yuan, M. L. and Nee, A. Y. C.},
  date = {2008},
  journaltitle = {International Journal of Production Research},
  volume = {46},
  pages = {2707--2742},
  issn = {0020-7543},
  doi = {10/dcq7zd},
  url = {https://doi.org/10.1080/00207540601064773},
  urldate = {2020-02-07},
  number = {10}
}

@report{oskoeiSurveyEdgeDetection2010,
  title = {A {{Survey}} on {{Edge Detection Methods}}},
  author = {Oskoei, Mohammadreza Asghari and Hu, Huosheng},
  date = {2010},
  pages = {36},
  institution = {{University of Essex}},
  langid = {english},
  number = {CES-506},
  type = {Technical Report}
}

@report{sobelCameraModelsMachine1972,
  title = {Camera {{Models}} and {{Machine Perception}}},
  author = {Sobel, Irwin},
  date = {1972},
  pages = {60},
  langid = {english},
  number = {CS0016},
  series = {},
  type = {Technion Technical Report}
}

@article{soriaDenseExtremeInception2020,
  title = {Dense {{Extreme Inception Network}}: {{Towards}} a {{Robust CNN Model}} for {{Edge Detection}}},
  shorttitle = {Dense {{Extreme Inception Network}}},
  author = {Soria, Xavier and Riba, Edgar and Sappa, Angel D.},
  date = {2020},
  url = {http://arxiv.org/abs/1909.01955},
  urldate = {2020-02-07},
  archivePrefix = {arXiv},
  eprint = {1909.01955},
  eprinttype = {arxiv},
}

@inproceedings{xieHolisticallyNestedEdgeDetection2015,
  title = {Holistically-{{Nested Edge Detection}}},
  booktitle = {{{IEEE}} {{International Conference}} on {{Computer Vision}}},
  author = {Xie, Saining and Tu, Zhuowen},
  date = {2015},
  pages = {1395--1403},
  url = {http://openaccess.thecvf.com/content_iccv_2015/html/Xie_Holistically-Nested_Edge_Detection_ICCV_2015_paper.html},
  urldate = {2019-12-08},
  eventtitle = {Proceedings of the {{IEEE International Conference}} on {{Computer Vision}}},
}

@inproceedings{yangRecovering3DPlanes2018,
  title = {Recovering {{3D Planes}} from a {{Single Image}} via {{Convolutional Neural Networks}}},
  booktitle = {{{European Conf.}} on {{Computer Vision}}},
  author = {Yang, Fengting and Zhou, Zihan},
  date = {2018},
  pages = {85--100},
  url = {http://openaccess.thecvf.com/content_ECCV_2018/html/Fengting_Yang_Recovering_3D_Planes_ECCV_2018_paper.html},
  urldate = {2020-02-16},
  eventtitle = {Proceedings of the {{European Conference}} on {{Computer Vision}} },
}

@inproceedings{yuUsefulVisualizationTechnique2010,
  title = {A {{Useful Visualization Technique}}: {{A Literature Review}} for {{Augmented Reality}} and Its {{Application}}, Limitation \& Future Direction},
  shorttitle = {A {{Useful Visualization Technique}}},
  booktitle = {Visual {{Information Communication}}},
  author = {Yu, Donggang and Jin, Jesse Sheng and Luo, Suhuai and Lai, Wei and Huang, Qingming},
  editor = {Huang, Mao Lin and Nguyen, Quang Vinh and Zhang, Kang},
  date = {2010},
  pages = {311--337},
  publisher = {{Springer US}},
  location = {{Boston, MA}},
  doi = {10/bfs7jb},
  isbn = {978-1-4419-0312-9},
  langid = {english},
}
\end{document}